%% file: main.tex
\documentclass[letterpaper, 10 pt, conference, twoside]{ieeeconf}  %

\IEEEoverridecommandlockouts                              %

\input{preamble}
\input{acronyms}
\input{shortcuts.tex}

\title{\LARGE \bf
\varTitle: Rasterization-based Path Planning \\
in 2D Gaussian Splatting Map
}

\author{Jiwon Park$^{1}$, Dong-Uk Seo$^{2}$ and Hyun Myung$^{2*}$
    \thanks{$^*$Corresponding author: Hyun Myung.}%
}

\begin{document}
\maketitle

\IEEEpeerreviewmaketitle

\input{abstract}
\input{introduction}

\input{related_work}
\input{preliminaries.tex}
\input{methodology}
\input{experiments}

\input{conclusion}

\bibliographystyle{IEEEtran}
\bibliography{abrv,reference}

\end{document}

%% file: preamble.tex
\makeatletter
\let\NAT@parse\undefined
\makeatother
\usepackage[T1]{fontenc}
\usepackage{cite}
\usepackage{amssymb,amsfonts}
\usepackage{hyperref}
\usepackage{algorithm}
\usepackage{algorithmicx}
\usepackage{algpseudocode}
\usepackage{graphicx}
\usepackage{textcomp}
\usepackage{mathtools}
\usepackage{changes}
\usepackage[font=footnotesize,labelformat=simple]{subcaption}
\usepackage{xcolor}
\usepackage{color, soul}
\usepackage{colortbl}   
\usepackage{transparent}
\usepackage{amsmath}
\usepackage{booktabs}
\usepackage{tabularx}
\usepackage{multirow}
\usepackage{xstring}
\usepackage{hhline}
\usepackage{kotex}
\usepackage{siunitx}
\usepackage{comment}
\usepackage{physics}
\usepackage{tikz}
\usepackage{cleveref}
\usepackage{gensymb}
\usepackage{threeparttable}
\usepackage{caption}
\usepackage{bm}
\usepackage{sjmacros}
\usepackage{adjustbox}
\usepackage{acro}
\usepackage{siunitx}    %
\usepackage[font=small]{caption}
\usepackage{xspace}

\PassOptionsToPackage{table}{xcolor}
\definecolor{oursrow}{RGB}{226,241,239}

\def\thesubsection{\thesection.\Alph{subsection}}

%% file: acronyms.tex
\DeclareAcronym{GS}{
  short = GS,
  long  = Gaussian splatting,
  short-indefinite = a,
  long-indefinite  = a
}

\DeclareAcronym{2DGS}{
  short = 2DGS,
  long  = 2D Gaussian splatting,
  short-indefinite = a,
  long-indefinite  = a
}

\DeclareAcronym{NeRF}{
  short = NeRF,
  long  = Neural Radiance Field,
  short-indefinite = a,
  long-indefinite  = a
}

\DeclareAcronym{CFM}{
  short = CFM,
  long  = conditional flow matching,
  short-indefinite = a,
  long-indefinite = a
}

\DeclareAcronym{RL}{
  short = RL,
  long  = reinforcement learning,
  short-indefinite = an,
  long-indefinite = a
}

\DeclareAcronym{DRL}{
  short = DRL,
  long  = deep reinforcement learning,
  short-indefinite = a,
  long-indefinite = a
}

\DeclareAcronym{FM}{
  short = FM,
  long  = flow matching,
  short-indefinite = an,
  long-indefinite = a
}

\DeclareAcronym{ODE}{
  short = ODE,
  long = ordinary differential equations,
  short-indefinite = an,
  long-indefinite = an
}

\DeclareAcronym{CNF}{
   short = CNF,
   long = continuous normalizing flows,
   short-indefinite = a,
   long-indefinite = a
}

\DeclareAcronym{PPO}{
  short = PPO,
  long  = proximal policy optimization,
  short-indefinite = a,
  long-indefinite = a
}

\DeclareAcronym{VAE}{
  short = VAE,
  long  = variational autoencoder,
  short-indefinite = a,
  long-indefinite = a
}

\DeclareAcronym{CENet}{
  short = CENet,
  long  = context-aided estimator network,
  short-indefinite = a,
  long-indefinite = a
}

\DeclareAcronym{IMU}{
  short = IMU,
  long  = inertial measurement unit,
  short-indefinite = an,
  long-indefinite = an
}

\DeclareAcronym{ID}{
  short = ID,
  long  = in-distribution,
  short-indefinite = an,
  long-indefinite = an
}

\DeclareAcronym{OOD}{
  short = OOD,
  long  = out-of-distribution,
  short-indefinite = an,
  long-indefinite = an
}

\DeclareAcronym{MLP}{
  short = MLP,
  long  = multi-layer perceptron,
  short-indefinite = an,
  long-indefinite = a
}
\DeclareAcronym{3DGS}{
  short = 3DGS,
  long  = 3D Gaussian splatting,
  short-indefinite = a,
  long-indefinite  = a
}

\DeclareAcronym{LCC}{
  short = LCC,
  long  = local cylindrical coordinate,
  short-indefinite = an,
  long-indefinite  = a
}

\DeclareAcronym{ESDF}{
  short = ESDF,
  long  = Euclidean signed distance field,
  short-indefinite = an,
  long-indefinite  = a
}

\DeclareAcronym{UGV}{
  short = UGV,
  long  = unmanned ground vehicle,
  short-indefinite = a,
  long-indefinite  = an
}

\DeclareAcronym{PRM}{
  short = PRM,
  long  = probabilistic roadmap,
  short-indefinite = a,
  long-indefinite  = a
}

\DeclareAcronym{GT}{
  short = GT,
  long  = ground truth,
  short-indefinite = a,
  long-indefinite  = a
}

\DeclareAcronym{SPL}{
  short = SPL,
  long  = success weighted by path length,
  short-indefinite = an,
  long-indefinite  = a
}

%% file: shortcuts.tex
\newcommand{\varTitle}{2DGS-Planner\xspace}

\newcommand{\scene}[1]{\texttt{#1}} 

%% file: abstract.tex
\begin{abstract}
Gaussian splatting provides an explicit and efficiently rasterizable scene representation for robot navigation. 
However, individual Gaussian primitives may not reliably represent obstacles as they are jointly optimized through alpha-composited rendering from a finite set of reconstruction views. 
We propose 2DGS-Planner, a path planner for ground robots that reads planning-relevant geometry from a 2D Gaussian splatting (2DGS) map through rasterization, rather than treating individual Gaussian primitives as obstacles.
During offline roadmap construction, multi-view attribution converts rendered normal dispersion into structural scores for non-ground disks supported by the reconstruction views. 
These scores guide adaptive node sampling on the ground. 
Path-aligned orthographic queries validate candidate edges, while cylindrical queries estimate local clearance fields that are cached on the edges. 
During online planning, graph search initializes a route, and path refinement reuses the cached fields while accounting for the robot’s dimensions and ground constraints. 
Experiments demonstrate improved roadmap connectivity, more accurate clearance estimation, and higher planning success compared with the tested baselines.
These results support rasterization as an effective geometric query interface for planning directly on Gaussian maps.
Code and data are available at \href{https://2dgs-planner.github.io/}{https://2dgs-planner.github.io/}.
\end{abstract}

%% file: introduction.tex
\section{Introduction}
\label{sec:introduction}

Autonomous navigation requires spatial reasoning about the environment to find safe and efficient paths. 
This reasoning is closely tied to how the environment is represented, since planners interact with a map through the spatial queries it can support efficiently.
Occupancy grids enable graph-based search through discrete occupancy lookups~\cite{elfes1989occupancy}, while signed distance fields provide differentiable distance queries for gradient-based trajectory optimization~\cite{ratliff2009chomp, oleynikova2017voxblox}.
Point-based geometric maps can be queried directly for collision checking in sampling-based planning~\cite{karaman2011sampling}.

\input{fig_01.tex}

Recently, learned scene representations have been explored as environment maps for navigation~\cite{adamkiewicz2022vision, kwon2023renderable, dai2026neural, splatnav, splatblox, gsplanner}.
\ac{NeRF}-based approaches obtain collision evidence directly from the learned density~\cite{adamkiewicz2022vision}. 
Other methods rely on rendered outputs for localization, image-goal navigation, and traversability-aware planning~\cite{kwon2023renderable, dai2026neural}.
Both approaches rely on evaluating an implicit field, which can be computationally expensive for real-time planning.

\ac{GS}~\cite{3dgs} provides an explicit alternative, encoding scenes with Gaussian primitives and rendering images from arbitrary viewpoints through differentiable rasterization.
Existing \ac{GS}-based planners generally treat the primitives directly as obstacles or convert the map into an intermediate geometric representation~\cite{splatnav, splatblox, gsplanner}.
However, visually plausible renderings do not guarantee that the underlying primitives align with the composited surface (Fig.~\ref{fig:primitive_reliability}).
Individual primitives may therefore fail to reliably represent scene geometry.

Two properties of GS training motivate rasterization-based geometric queries.
First, the primitives are jointly optimized through their contributions to alpha-composited renderings, so the training objective constrains the composited geometry rather than any individual primitive.
This motivates querying the composited geometry instead of treating primitives as independent obstacles.
Second, this supervision comes from only a finite set of posed training views.
Geometry outside their observation support is therefore less constrained by the training data.
A planner should consider not only the rendered surface geometry but also how well that surface was observed during training.

We therefore propose \varTitle~(Fig.~\ref{fig:teaser}), a ground-robot path planner that treats rasterization as the geometric query interface to a \ac{GS} map.
We build on \ac{2DGS}~\cite{2dgs}, whose training aligns each disk with the local surface through a normal consistency term, so that surface normals become a rendered channel alongside depth and alpha.
The framework constructs a roadmap directly from rasterized geometric queries without converting the scene into an intermediate geometric representation. 
During offline roadmap construction, multi-view attribution derives structural scores from rendered normal dispersion. 
Scores from non-ground disks supported by the reconstruction views guide adaptive node sampling on the ground, placing more nodes near corners and cluttered areas and fewer in open regions. 
Path-aligned orthographic queries validate candidate connections, while cylindrical queries estimate local clearance fields that are cached on retained edges. 
During online planning, graph search finds an initial path, which is then refined using the cached fields while accounting for the robot’s dimensions and ground constraints.

The main contributions are as follows:
\begin{itemize}
\item We propose \varTitle, a ground-robot path planner that queries 2DGS maps through rasterization without an intermediate geometric representation.
\item We introduce multi-view attribution that converts rendered normal dispersion into structural scores for adaptive roadmap sampling.
\item We formulate path-aligned orthographic and cylindrical queries for edge validation and local clearance estimation, with the clearance fields cached for path refinement.
\item We evaluate the proposed framework from geometric reasoning to complete planning, including structural attribution, clearance estimation, end-to-end path planning, and real-world robot experiments.
\end{itemize}

%% file: fig_01.tex
\begin{figure}[!t]
    \captionsetup{font=footnotesize}
    \centering
    \includegraphics[width=\columnwidth]{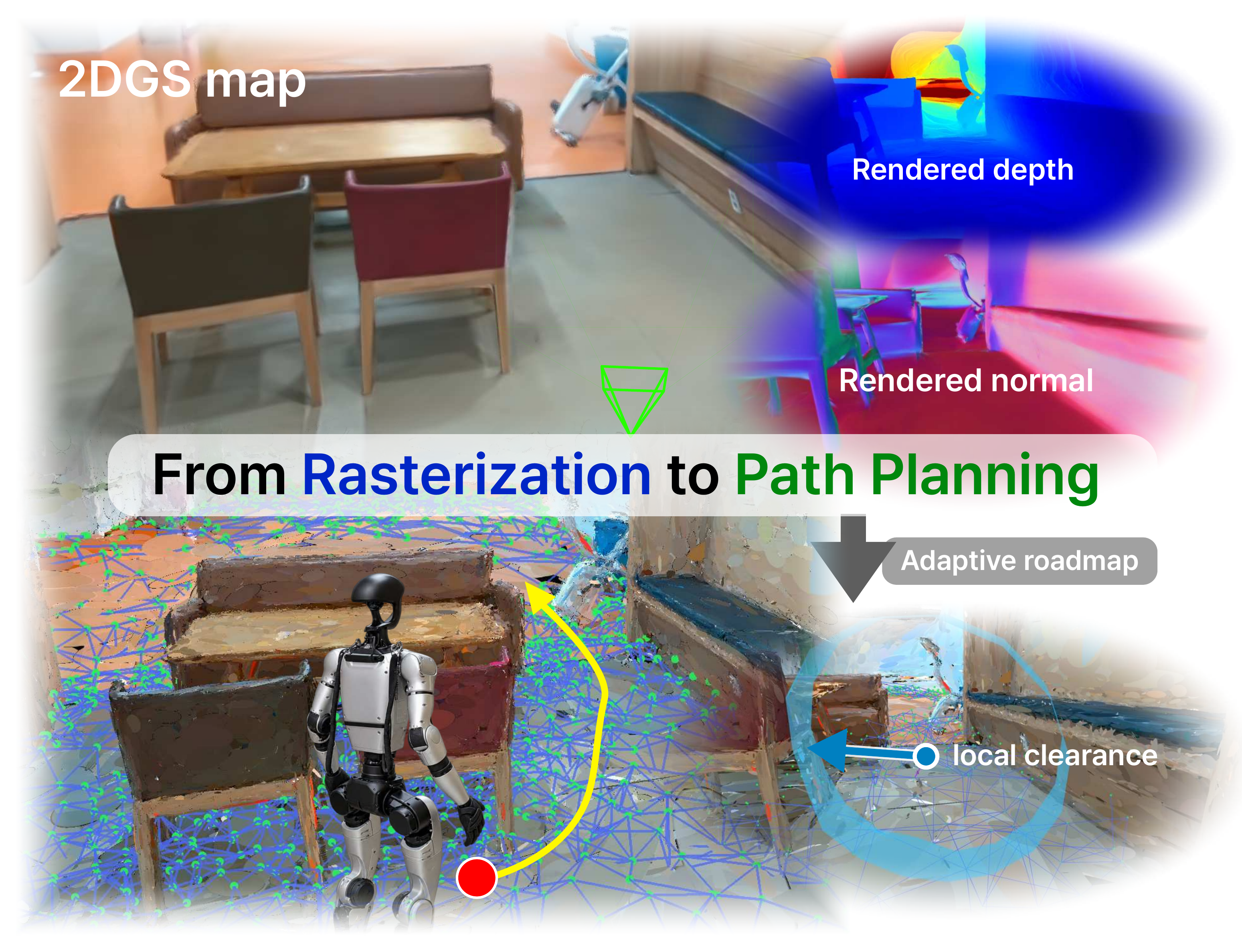}
    \caption{
    \varTitle uses rasterization to read surface structure and local clearance directly from a \ac{2DGS} map.
    Multi-view normal dispersion guides structure-aware roadmap construction, while path-aligned rasterization queries validate candidate motions and provide local clearance for path refinement. 
    The resulting roadmap and refined path are shown in the reconstructed scene.
    }
    \label{fig:teaser}
\end{figure}

%% file: related_work.tex
\section{Related Work}\label{sec:related_work}

\input{a.gs_rendering}
\input{b.gs_planning}

\input{fig_02.tex}

\input{fig_06.tex}

%% file: a.gs_rendering.tex
\subsection{Rasterization-based Geometry in Gaussian Splatting}
\label{rw:rendering}

\ac{3DGS}~\cite{3dgs} represents a scene with volumetric Gaussian
primitives, while \ac{2DGS}~\cite{2dgs} replaces them with planar
Gaussian disks to impose a surface-oriented geometric representation.
During rasterization, the \ac{2DGS} renderer computes ray--disk
intersections and alpha-composites the intersected disks along each ray, yielding geometric quantities such as depth and surface normals in 3D.
Recent methods have increasingly exploited such rasterized quantities
as geometric signals.
PGSR~\cite{pgsr} renders plane distances and normals for surface
reconstruction, while RaDe-GS~\cite{radegs} derives depth and surface
normals directly from rasterized 3D Gaussians.
3DGUT~\cite{wu2025threedgut} further extends Gaussian rasterization
beyond standard pinhole projection to support secondary rays.
These methods use rasterization primarily for rendering or reconstruction, while we use it as a geometric query interface for path planning on a fixed \ac{2DGS} map.

%% file: b.gs_planning.tex
\subsection{Planning on Gaussian Splatting}
\label{rw:planning}

GS-based motion planning and control differ in how they obtain collision information.
Splat-Nav~\cite{splatnav} constructs safe corridors from Gaussian ellipsoids, while SAFER-Splat~\cite{safersplat} formulates control barrier functions over the primitives.
SPLANNING~\cite{splanning} derives collision-probability bounds from a normalized Gaussian field, and
FOCI~\cite{foci} optimizes trajectories using overlap between Gaussian representations of the robot and the scene, including trajectories for a legged ground robot.

Other approaches use auxiliary geometric representations.
Splatblox~\cite{splatblox} supports wheeled and legged outdoor navigation through a traversability-aware \ac{ESDF}, while LiftNav~\cite{liftnav} uses a joint TSDF--GS map for semantic navigation and trajectory
optimization.
SplatlessDF~\cite{splatlessdf} couples \ac{2DGS} with a separately parameterized continuous distance field
that supplies distance and gradient queries for planning.

Rendering has also been used directly for collision avoidance.
ReMoSPLAT~\cite{remosplat} renders depth from virtual cameras attached to a mobile manipulator and uses the resulting distances for reactive control.
Whereas ReMoSPLAT targets reactive collision avoidance, we use rasterization for ground-robot path planning on a fixed \ac{2DGS} map, with queries organized around candidate ground motions to evaluate and retain clearance for subsequent refinement.

%% file: fig_02.tex
\begin{figure}[!t]
	\captionsetup{font=footnotesize}
	\centering
	\includegraphics[width=1.0\columnwidth]{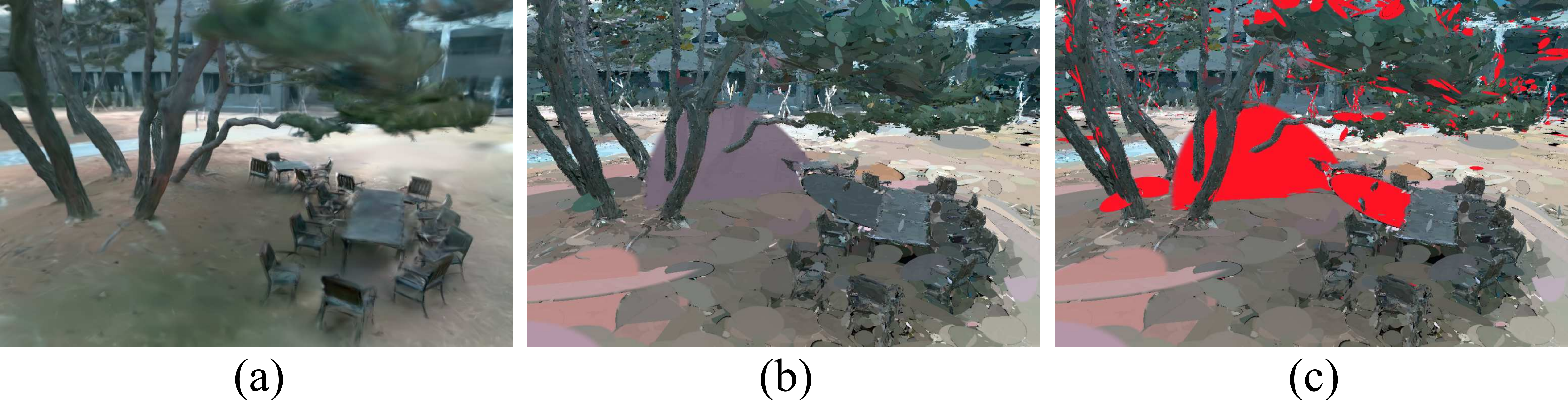}
    \caption{
    Geometric discrepancy between individual 2DGS primitives and the alpha-composited surface. 
    (a) RGB rendering. 
    (b) Underlying Gaussian disks visualized at their \(1\sigma\) extent. 
    (c) Disks with large depth discrepancies from the rendered surface, highlighted in red. Visually consistent rendering can therefore coexist with inaccurate primitive-level geometry.
    }
	\label{fig:primitive_reliability}
    \vspace{-0.2cm}
\end{figure}

%% file: fig_06.tex
\begin{figure*}[!t]
	\captionsetup{font=footnotesize}
	\centering 
	\includegraphics[width=1.0\textwidth]{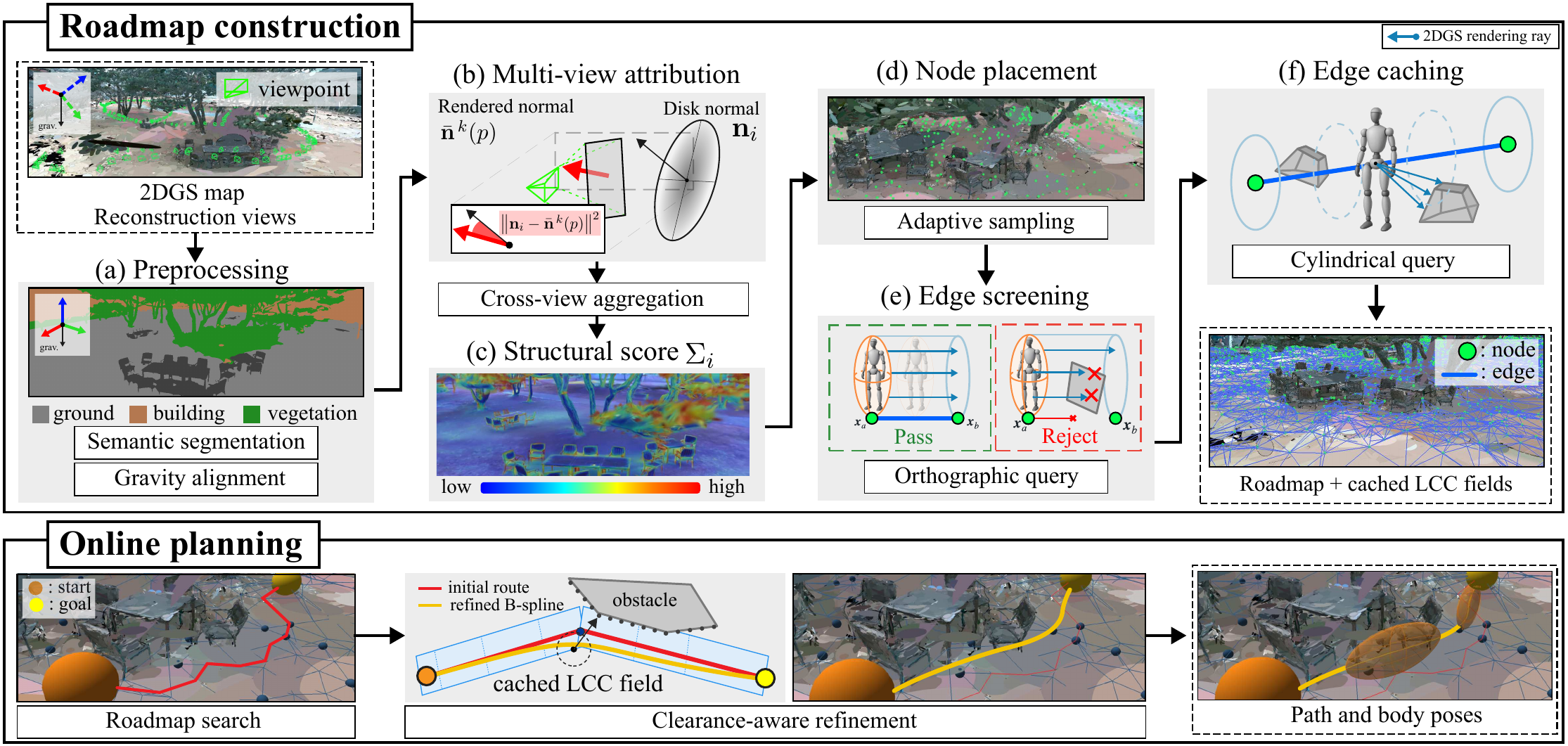}
    \caption{
    Overview of \varTitle{}.
    \textbf{Roadmap construction (top):}
    Given a 2DGS map and its reconstruction views,
    (a) semantic preprocessing identifies supporting ground and aligns the map with gravity.
    (b)--(c) Multi-view attribution aggregates contribution-weighted
    per-pixel normal dispersion across reconstruction views to obtain
    structural scores $\Sigma_i$ for observation-supported non-ground disks.
    (d) The scores guide adaptive node placement.
    (e) Orthographic queries screen candidate edges, and
    (f) cylindrical queries estimate surrounding clearance and cache
    the resulting fields on retained edges.
    \textbf{Online planning (bottom):}
    Roadmap search using $A^*$ initializes a route, which is refined
    as a B-spline using surface samples back-projected from the cached
    clearance fields while accounting for robot dimensions and ground
    constraints.
    The final panel shows the refined path and its corresponding
    gravity-aligned body poses, with headings following the path tangents.
    }
	\label{fig:framework}
    \vspace{-0.2cm}
\end{figure*}

%% file: preliminaries.tex
\section{Preliminaries}
\label{sec:prelim}

A 2DGS map consists of $N$ oriented planar Gaussian disks $\mathcal{G}=\{G_i\}_{i=1}^{N}$ reconstructed from $K$ posed images.
Their camera poses form the reconstruction views $\mathcal{V}=\{T_k\}_{k=1}^{K}$.
Each disk $G_i$ has a center $\boldsymbol{\mu}_i\in\mathbb{R}^3$, orthonormal tangent axes $\mathbf{t}_i^u,\mathbf{t}_i^v$,
positive scales $s_i^u,s_i^v$, and opacity $\alpha_i\in[0,1]$.
Its unit normal is $\mathbf{n}_i=\mathbf{t}_i^u\times\mathbf{t}_i^v$.

A rasterization query consists of rays indexed by a query sample $p$, with origin $\mathbf{o}(p)\in\mathbb{R}^3$ and unit direction $\mathbf{r}(p)\in\mathbb{S}^2$.
Closed-form ray--disk intersections~\cite{2dgs} are evaluated in world space to obtain the Gaussian response $G_i(p)\in[0,1]$ and the hit distance $d_i(p)$ from the ray origin to the plane of each intersected disk.

Front-to-back alpha composition determines the contribution $w_i(p)$ of disk $G_i$ to sample $p$ and the accumulated opacity $A(p)$:
\begin{equation}
\begin{aligned}
w_i(p)
&=
\alpha_i G_i(p)
\prod_{j\prec_p i}
\left(1-\alpha_j G_j(p)\right),\\
A(p)
&=
\sum_i w_i(p)\in[0,1],
\end{aligned}
\label{eq:compositing}
\end{equation}
where $j\prec_p i$ indicates that the ray intersects disk $G_j$ before disk $G_i$.
For $A(p)>0$, the normalized contribution is $\hat{w}_i(p)=w_i(p)/A(p)$.

For a reconstruction view $T_k$, superscript $k$ denotes view-specific rendered quantities, such as $w_i^k(p)$ and $A^k(p)$; quantities without the superscript refer to the current query.

A ray is considered observed when $A(p)\ge\tau_A$, with $\tau_A=0.5$.
For observed rays, the rendered median depth $d^{\mathrm{med}}(p)$ is defined as the first hit distance at which the accumulated opacity reaches $0.5$.
Because low accumulated opacity may result from either empty space or insufficient observation, unobserved rays are treated as lacking geometric evidence rather than as free space.

%% file: methodology.tex
\section{\varTitle}\label{sec:method}

\input{a.overview}
\input{b.map_preprocessing}

\input{c.multi_view_attribution}
\input{e.roadmap_construction}

\input{d.motion_query}
\input{fig_04.tex}

\input{fig_05.tex}
\input{f.path_pose_optimization}

%% file: a.overview.tex
\subsection{Problem Formulation and Overview}
\label{sec:problem}

We plan for a ground robot on a static \ac{2DGS} map $\mathcal{G}$ with reconstruction views $\mathcal{V}$.
The robot body is modeled as a gravity-aligned ellipsoid with semi-axes $(a,b,c)$, whose center is maintained at a prescribed height $h$ above the supporting ground.
Given start and goal locations on the observed ground, we seek a connecting path that maintains body clearance and ground support.

Fig.~\ref{fig:framework} summarizes the overall pipeline.
The offline stage constructs a roadmap $\mathcal{H}=(\mathcal{N},\mathcal{E})$ over candidate robot positions.
Multi-view attribution determines the sampling density from rendered structural cues, while path-aligned queries validate candidate edges and cache local clearance on retained edges.
During online planning, A$^*$ initializes a graph route, which is subsequently refined using the cached clearance under body-clearance and ground constraints.

%% file: b.map_preprocessing.tex
\subsection{Map Preprocessing}
\label{sec:preprocess}

Map preprocessing assigns semantic labels to the Gaussian disks and aligns the reconstruction with gravity.
Semantic predictions from the reconstruction views are transferred to the disks to identify ground regions.
Ground-labeled disks satisfying the orientation and height criteria form $\mathcal{G}^{\mathrm{grd}}\subset\mathcal{G}$, which defines the support surface for roadmap sampling and path refinement.
All subsequent planning is performed in the gravity-aligned frame.

%% file: c.multi_view_attribution.tex
\subsection{Multi-View Attribution}
\label{sec:attribution}

Multi-view attribution estimates local structural complexity from the variation of rendered normals and uses it to guide roadmap sampling.

\paragraph{Per-pixel normal dispersion}
For a pixel $p$ in reconstruction view $T_k$ with $A^k(p)>0$, we compute the
contribution-weighted mean normal as follows:
\begin{equation}
\bar{\mathbf{n}}^{\,k}(p) = \sum_i \hat{w}_i^k(p)\mathbf{n}_i .
\end{equation}
Because $\|\mathbf{n}_i\|=1$, the corresponding dispersion is
\begin{equation}
\begin{aligned}
V^k(p) &= \sum_i \hat{w}_i^k(p)
\left\|\mathbf{n}_i-\bar{\mathbf{n}}^{\,k}(p)\right\|^2 \\
&= 1-\left\|\bar{\mathbf{n}}^{\,k}(p)\right\|^2 .
\end{aligned}
\label{eq:normal_dispersion}
\end{equation}
The mean normal is not renormalized, so its magnitude reflects agreement among the
contributing normals.
Dispersion therefore rises where multiple surface orientations contribute to a single
pixel, as occurs near corners, thin structures, and cluttered regions.

\paragraph{Per-disk attribution}
We aggregate the rendered dispersion across reconstruction views using each disk's
alpha-compositing contribution as follows~\cite{rotabulo2024revising}:
\begin{equation}
\nu_i = \frac{\sum_{k,p} w_i^k(p)V^k(p)}{\sum_{k,p} w_i^k(p)} .
\label{eq:disk_attribution}
\end{equation}
Observations in which a disk contributes more strongly therefore have greater influence
on its attributed dispersion.

\paragraph{Structural score}
We convert $\nu_i$ into a normalized structural score $\Sigma_i\in[0,1]$ using its
empirical percentile among eligible non-ground disks.
Eligibility requires sufficient total rendering contribution and support from a minimum
number of reconstruction views, with thresholds given in Sec.\,\ref{sec:impl}.
For each eligible disk, $\Sigma_i$ is the fraction of eligible disks with attributed
dispersion no greater than $\nu_i$, and ineligible disks are assigned $\Sigma_i=0$.
Ground disks are excluded from the ranking to prevent their numerous low-dispersion
values from biasing non-ground scores upward.
The resulting scores guide the sampling density of nearby ground candidates.

%% file: e.roadmap_construction.tex
\subsection{Roadmap Construction}
\label{sec:roadmap}
We construct a roadmap by sampling robot positions on the supporting ground and
connecting them through collision-free edges.

\paragraph{Nodes}
Candidate positions are sampled from ground-disk centers and lifted to the prescribed
ride height.
Each candidate $\mathbf{q}$ is assigned a local complexity score
\begin{equation}
u(\mathbf{q})=
\frac{1}{K_{\mathrm{nb}}}
\sum_{j\in\mathcal{I}(\mathbf{q})}
\left[1-\frac{\|\mathbf{q}-\boldsymbol{\mu}_j\|}{R}\right]_+
\Sigma_j ,
\label{eq:local_complexity}
\end{equation}
where $\mathcal{I}(\mathbf{q})$ contains the $K_{\mathrm{nb}}$ nearest non-ground disks,
$R$ is the neighborhood radius, and $[x]_+=\max(x,0)$.
The fixed denominator $K_{\mathrm{nb}}$ keeps the score low for candidates with few
contributing neighbors.
The percentile rank of $u(\mathbf{q})$ determines the target node spacing between
$\ell_{\max}$ and $\ell_{\min}$, assigning denser sampling to structurally complex
regions, while candidates with no non-ground disk within $R$ use $\ell_{\max}$.
Adaptive Poisson-disk thinning then produces the final node set
(Fig.~\ref{fig:roadmap_score}).

\paragraph{Edges}
A Gabriel graph~\cite{gabriel} proposes candidate connections between sampled nodes.
Each candidate edge is first screened by an orthographic query, after which a cylindrical
query estimates local clearance for the remaining edges.
Only edges satisfying both criteria are retained in $\mathcal{E}$ together with their
cached clearance fields for online path refinement.

%% file: d.motion_query.tex
\subsection{Path-Aligned Motion Queries}
\label{sec:motion_queries}
Both queries use a frame aligned with the candidate edge, with the orthographic query
screening motion along the edge and the cylindrical query recording surrounding
clearance for path refinement.

\paragraph{Edge-aligned frame}
An edge $e$ connects nodes at $\mathbf{x}_1,\mathbf{x}_2\in\mathbb{R}^3$, with length
$L=\|\mathbf{x}_2-\mathbf{x}_1\|$ and unit direction
$\mathbf{t}_e=(\mathbf{x}_2-\mathbf{x}_1)/L$.
With $\hat{\mathbf{z}}$ denoting the world up axis, the transverse axes are
\[
\mathbf{b}_1=
\frac{\hat{\mathbf{z}}\times\mathbf{t}_e}{\|\hat{\mathbf{z}}\times\mathbf{t}_e\|},
\qquad
\mathbf{b}_2=\mathbf{t}_e\times\mathbf{b}_1,
\]
which span the cross-sectional plane shared by both queries.

\paragraph{Orthographic footprint validation}
We sample $(u,v)$ on a regular grid in the $\mathbf{b}_1$--$\mathbf{b}_2$ plane and
retain samples satisfying $(u/(b+\delta))^2+(v/(c+\delta))^2\leq1$, where $\delta$ is
the safety margin.
Each sample casts a parallel ray from
$\mathbf{o}(u,v)=\mathbf{x}_1+u\mathbf{b}_1+v\mathbf{b}_2$ along $\mathbf{t}_e$.
The edge is rejected if any sampled ray observes a surface before reaching the
endpoint, that is, $A(u,v)\geq\tau_A$ and $d^{\mathrm{med}}(u,v)<L$.
This test screens translation between the two nodes but covers neither the
longitudinal overhang $a$ beyond either endpoint nor in-place rotations at a node.

\paragraph{Local cylindrical clearance}
For edges passing the orthographic screen, the cylindrical query records surrounding
geometry in a local cylindrical coordinate (LCC) system (Fig.~\ref{fig:lcc}),
parameterized by axial position $s\in[0,L]$ and azimuth $\theta\in(-\pi,\pi]$ measured
from $\mathbf{b}_1$ toward $\mathbf{b}_2$.
At regularly sampled pairs $(s,\theta)$, the radial rays are
\begin{equation}
\begin{aligned}
\mathbf{o}(s,\theta)&=\mathbf{x}_1+s\mathbf{t}_e,\\
\mathbf{r}(s,\theta)&=\cos\theta\,\mathbf{b}_1+\sin\theta\,\mathbf{b}_2 .
\end{aligned}
\label{eq:cylindrical_rays}
\end{equation}
The clearance field $\rho_e(s,\theta)=d^{\mathrm{med}}(s,\theta)$ is defined only for
observed rays with $A(s,\theta)\geq\tau_A$.
A slice at fixed $s$ is rejected if none of its rays is observed or if its minimum
observed clearance falls below the lateral half-width $b$, and the candidate edge is
rejected if any sampled slice fails.
Each retained edge stores $\rho_e$ for clearance queries during path refinement.

%% file: fig_04.tex
\begin{figure}[!t]
	\captionsetup{font=footnotesize}
	\centering
	\includegraphics[width=1.0\columnwidth]{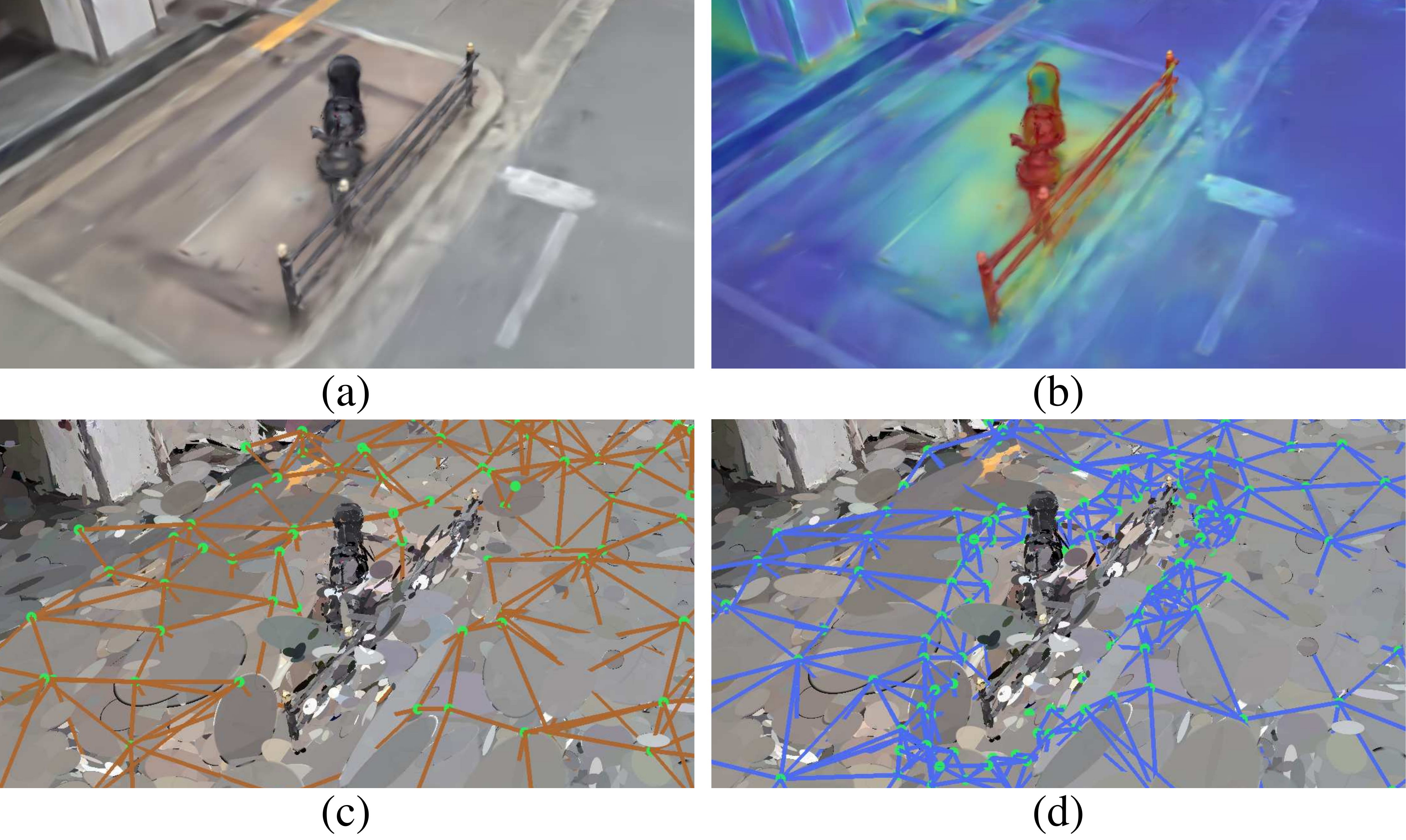}
    \caption{Structure-aware roadmap sampling.
    (a) RGB rendering.
    (b) Structural score produced by multi-view attribution.
    (c) Uniform sampling with its resulting roadmap.
    (d) Adaptive sampling using the structural scores, which concentrates nodes near locally complex structures while reducing sampling in open regions under the same node budget.
    }
	\label{fig:roadmap_score}
    \vspace{-0.2cm}
\end{figure}

%% file: fig_05.tex
\begin{figure}[!tb]
	\captionsetup{font=footnotesize}
	\centering
	\includegraphics[width=1.0\columnwidth]{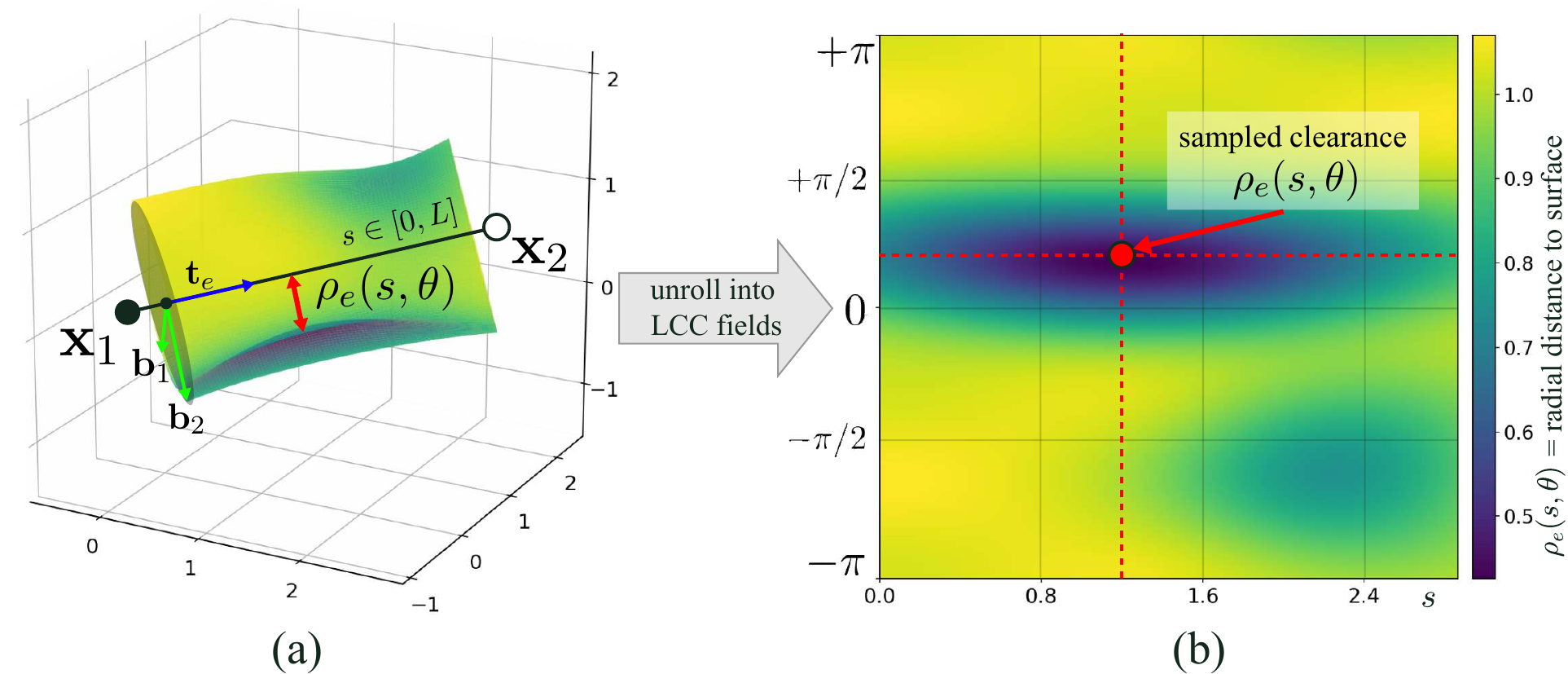}
    \caption{
    Local cylindrical clearance (LCC) representation.
    (a) For an edge \(e=(\mathbf{x}_1,\mathbf{x}_2)\), radial rasterization queries are cast from axial position \(s\) in the edge-aligned frame \(\{\mathbf{t}_e,\mathbf{b}_1,\mathbf{b}_2\}\). The resulting distance \(\rho_e(s,\theta)\) measures radial clearance to the observed surface at azimuth \(\theta\). 
    (b) Unrolling the measurements over \((s,\theta)\) forms the LCC field, which is cached on the edge and reused during path refinement. Colors encode radial clearance.
    }
	\label{fig:lcc}
    \vspace{-0.2cm}
\end{figure}

%% file: f.path_pose_optimization.tex
\subsection{Path Optimization}
\label{sec:path_optimization}

\paragraph{Initial path}
The start and goal are connected to nearby roadmap nodes using the same path-aligned
queries, and A$^*$ provides a graph route for B-spline initialization.
The endpoints remain fixed during refinement.

\paragraph{Corridor-constrained refinement}
Each cached \ac{LCC} field is back-projected into local surface samples and reused
throughout refinement without additional rendering.
For each spline sample $\mathbf{q}_n$, clearance $c(\mathbf{q}_n)$ is evaluated as the
distance to the nearest surface sample associated with the corresponding route edge.
Unlike $\rho_e(s,\theta)$, which is measured radially from the original edge axis,
$c(\mathbf{q}_n)$ is evaluated at the current refined path position.
To account for the additional space required by an elongated body during turning, the
required clearance increases with the local path curvature $\kappa_n$ as
\begin{equation}
r_n=\min\!\left(
 b+\frac{1}{2}a^2\kappa_n,\,\max(a,b)
\right).
\label{eq:required_clearance}
\end{equation}
The spline control points are then updated iteratively by minimizing
\begin{equation}
\begin{aligned}
E={}&w_{\mathrm{len}}\sum_n
\|\mathbf{q}_{n+1}-\mathbf{q}_n\|^2\\
&+w_{\mathrm{clr}}\sum_n
\left[r_n+\delta-c(\mathbf{q}_n)\right]_+^2,
\end{aligned}
\label{eq:path_objective}
\end{equation}
where $w_{\mathrm{len}}$ and $w_{\mathrm{clr}}$ weight path regularization against the
clearance penalty, with control-point heights constrained to the prescribed ride height
above the estimated local ground.
The robot heading follows the path tangent.
The refined path is finally revalidated using the orthographic query, and failed spans
fall back to the corresponding validated graph segments.

%% file: experiments.tex
\section{Experiments}
\label{sec:experiments}

The experiments evaluate \varTitle{} at three levels: roadmap
construction, geometric query fidelity, and end-to-end planning.
Sec.\,\ref{sec:exp_roadmap} evaluates whether multi-view attribution
captures local structural complexity and improves adaptive roadmap
sampling.
Sec.\,\ref{sec:exp_clearance} evaluates the fidelity of cached
\ac{LCC} clearance fields against a voxelized representation of the
same rendered surfaces.
Sec.\,\ref{sec:exp_planning} evaluates the resulting end-to-end planning
performance, including ablations of adaptive sampling and
\ac{LCC}-based refinement.
Sec.\,\ref{sec:exp_robot_size} further illustrates how robot dimensions
affect roadmap connectivity and the resulting path through
robot-specific orthographic footprint queries.
Finally, Sec.\,\ref{sec:exp_real} demonstrates the planner on a physical
humanoid robot.

\input{fig_structure.tex} %
\input{a.setup.tex}
\input{c.roadmap.tex}
\input{b.clearance.tex}
\input{fig_clearance.tex} %
\input{tab_planning.tex} %
\input{fig_quantitative.tex} %
\input{d.planning.tex}
\input{fig_sweep_cost.tex}
\input{fig_cost.tex}
\input{fig_robot_size.tex} %
\input{f.qualitative.tex}
\input{e.real.tex}
\input{fig_real.tex}

%% file: fig_structure.tex
\begin{figure}[!tb]
\captionsetup{font=footnotesize}
\centering
\includegraphics[width=\columnwidth]{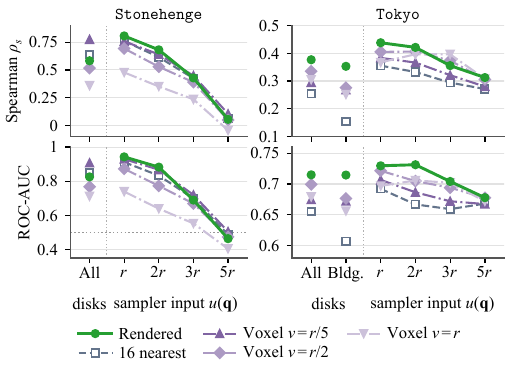}
\caption{
Correlation and discrimination performance of structural scores with respect to reference-mesh normal dispersion on \scene{Stonehenge} (exact mesh, left) and \scene{Tokyo} (proxy mesh, right). The top and bottom rows report Spearman correlation \(\rho_s\) and ROC-AUC, respectively. Disk-level results compare rendered attribution with geometric neighborhood scores, while sampler-input results evaluate the aggregated node score \(u(\mathbf{q})\) over reference radii \(r\)–\(5r\). For \scene{Tokyo}, “Bldg.” denotes the building subset. Dotted horizontal lines indicate chance-level ROC-AUC.
}    
\label{fig:structure}
\vspace{-0.2cm}
\end{figure}

%% file: a.setup.tex
\subsection{Evaluation Environments}
\label{sec:exp_env}

Quantitative evaluation is performed on two 2DGS scenes reconstructed
from posed RGB images.
\scene{Stonehenge} provides an exact synthetic reference mesh, whereas
\scene{Tokyo}~\cite{niantic_real2sim_2026} provides an aligned proxy mesh
for a partially observed real-world reconstruction.
Together, they enable evaluation under both exact and imperfect
reference geometry and are used for the experiments in
Secs.\,\ref{sec:exp_roadmap}--\ref{sec:exp_planning}.

\subsection{Baselines}
\label{sec:exp_baselines}

We compare against sampling-based planners and an \ac{ESDF}-based planner, all operating on the same \ac{2DGS} reconstruction.
Disk \ac{PRM}~\cite{kavraki1996probabilistic}, RRT$^*$~\cite{karaman2011sampling}, and Informed RRT$^*$~\cite{gammell2014informed} treat each Gaussian disk as an obstacle truncated at its $1\sigma$ elliptical extent and evaluate collisions using exact point-to-ellipse distances~\cite{eberly}.
For Disk \ac{PRM}, we evaluate uniformly sampled roadmaps with node budgets of $1$k, $10$k, and $60$k to examine how increased sampling affects roadmap construction cost and planning performance.
The ground-constrained \ac{ESDF} baseline fuses depth panoramas rendered from the reconstruction views into an occupancy grid, performs A$^*$ search on the resulting distance field, and applies B-spline refinement.
This baseline evaluates planning through an intermediate volumetric representation.

We evaluate uniform and adaptive roadmap sampling, each with and without \ac{LCC}-based refinement.

\subsection{Implementation Details}
\label{sec:impl}

The robot is modeled as a sphere throughout the experiments, with radius $r=0.05$ in scene units for the non-metric \scene{Stonehenge} map and $r=0.50\mathrm{m}$ for the metric \scene{Tokyo} map. The anisotropic robot-body model of Sec.\,\ref{sec:problem} is used only in the robot-size-aware planning experiment.
The roadmap uses node spacing $\ell\in[0.10,0.25]$ in the corresponding map units and a $64{\times}128$ cylindrical clearance field per edge. For structural attribution, disks are retained with support $\tau_S=10^{-3}$ from at least $K_{\min}=3$ views.
Path refinement uses a B-spline with 12 control points, maintains the prescribed ride height, and revalidates the resulting path using the orthographic query of Sec.\,\ref{sec:motion_queries}. Semantic labels are obtained with SegFormer-B5~\cite{xie2021segformer} pretrained on ADE20K~\cite{zhou2017ade20k} and grouped into task-relevant categories.
All experiments are conducted on an NVIDIA RTX 4090, an Intel Core i9-10900K, and $94\mathrm{GB}$ of RAM.

%% file: c.roadmap.tex
\subsection{Structure-Aware Roadmap Construction}
\label{sec:exp_roadmap}
We evaluate whether rendered attribution captures local scene structure well enough to serve as a sampling prior.
Fig.~\ref{fig:structure} compares rendered attribution (Sec.\,\ref{sec:attribution}) with disk-neighborhood scores on \scene{Stonehenge} and \scene{Tokyo}, taking normal dispersion on the reference mesh as the measure of local structural complexity.
Spearman correlation $\rho_s$ measures whether more complex regions receive higher scores, and ROC-AUC measures how well the scores separate the top $25\%$ from the bottom $50\%$ of reference dispersion.

At the disk level, the two scenes rank differently.
On \scene{Stonehenge}, the finest voxel score correlates most strongly with the exact mesh, indicating that individual disk geometry represents local surface structure well in a synthetic scene.
On \scene{Tokyo}, rendered attribution instead achieves the highest $\rho_s$ and ROC-AUC for both the full obstacle set and the building subset, since deriving structure from surfaces supported by visible disk contributions across views depends less on noisy
local geometry.
The geometric scores also lose consistency with the reference as the neighborhood scale grows, revealing their sensitivity to the chosen spatial scale.

For node placement, the relevant quantity is the aggregated score $u(\mathbf q)$ rather than individual disk accuracy.
With $R=3r$, rendered attribution achieves the highest $\rho_s$ and ROC-AUC at $R_{\mathrm{ref}}=r$ and $2r$ on both scenes, despite its lower disk-level ranking on \scene{Stonehenge}, so stronger primitive-level agreement does not necessarily yield a better sampling prior.
The ranking again becomes less consistent at larger scales, supporting rendered attribution as a local, robot-scale structural signal.
Whether this signal improves planning is evaluated directly in Sec.\,\ref{sec:exp_planning}, where uniform and attribution-guided sampling are compared
under a matched node budget.

%% file: b.clearance.tex
\subsection{Clearance Query Fidelity}
\label{sec:exp_clearance}

We compare cached \ac{LCC} fields with \ac{ESDF}s constructed from identical rendered \ac{2DGS} surface samples to isolate the effect of the clearance representation.
The exact \scene{Stonehenge} mesh provides reference clearance, and the evaluation focuses on configurations near the collision boundary, where clearance estimation errors are most relevant to planning decisions.

Fig.~\ref{fig:exp_clearance} shows that \ac{LCC} yields lower clearance error and fewer false-blocked configurations than the tested \ac{ESDF}s, better preserving free space supported by the reference geometry.
The higher false-blocked rates of the \ac{ESDF}s are consistent with volumetric occupancy fusion reducing the represented free space, whereas \ac{LCC} retains clearance in path-aligned fields without additional volumetric discretization.

These results indicate that \ac{LCC} more faithfully preserves the rendered surface information relevant to local feasibility decisions.
False-safe cases remain, however, so the estimated clearance does not provide a collision-free guarantee.
\ac{LCC} construction is also faster than that of the finer \ac{ESDF} grids.

%% file: fig_clearance.tex
\begin{figure}[!t]
\captionsetup{font=footnotesize}
\centering
\includegraphics[width=\columnwidth]{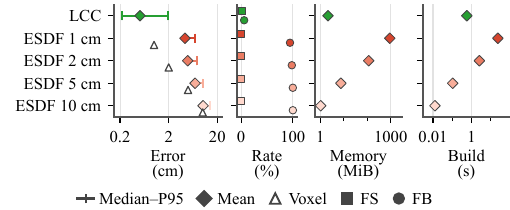}
\caption{
    Clearance fidelity and representation cost on \scene{Stonehenge}. LCC fields and ESDFs are constructed from the same rendered 2DGS surface samples and evaluated on 5,716 near-boundary queries satisfying \(\lvert c_{\mathrm{GT}}-r\rvert<5\,\mathrm{cm}\) along 30 paths. Clearance error is summarized by the mean (diamond) and median–95th percentile interval (bar); the remaining panels report false-safe (FS) and false-blocked (FB) rates, memory usage, and construction time. Error, memory, and time are shown on logarithmic scales.
    }
\label{fig:exp_clearance}
\vspace{-0.2cm}
\end{figure}

%% file: tab_planning.tex
\definecolor{oursgreen}{HTML}{2E8B57}
\newcommand{\best}{\cellcolor{oursgreen!55}}
\newcommand{\secondbest}{\cellcolor{oursgreen!18}}

\begin{table}[t]
\centering
\caption{
Reference-geometry evaluation on $50$ start--goal pairs per scene.
Collision pass reports paths satisfying the reference check at tolerance
$\varepsilon$, while SPL evaluates successes at
$\varepsilon=\varepsilon_{\mathrm{ref}}$ using
$\min(1,L_{\mathrm{ref}}/L)$, where $L_{\mathrm{ref}}$ and $L$ are the
reference and returned path lengths.
We use $\varepsilon_{\mathrm{ref}}=0.005$ for \scene{Stonehenge} and
$0.10\,\mathrm{m}$ for \scene{Tokyo}.
}
\label{tab:planning}

\scriptsize
\setlength{\tabcolsep}{0pt}
\begin{tabular*}{\columnwidth}{@{\extracolsep{\fill}}l rrr rrr@{}}
\toprule
& \multicolumn{3}{c}{\scene{Stonehenge}}
& \multicolumn{3}{c}{\scene{Tokyo}} \\
\cmidrule(lr){2-4}\cmidrule(lr){5-7}
Method
& $\varepsilon{=}\varepsilon_{\mathrm{ref}}\uparrow$
& $\varepsilon{=}0\uparrow$
& SPL$\uparrow$
& $\varepsilon{=}\varepsilon_{\mathrm{ref}}\uparrow$
& $\varepsilon{=}0\uparrow$
& SPL$\uparrow$ \\
\midrule
Disk \acs{PRM} ($1$k)        & 21/50 & 15/50 & 0.335 & 6/50  & 6/50  & 0.091 \\
Disk \acs{PRM} ($10$k)       & 43/50 & 29/50 & 0.746 & 10/50 & 10/50 & 0.176 \\
Disk \acs{PRM} ($60$k)       & 45/50 & 27/50 & 0.810 & 10/50 & 9/50  & 0.168 \\
RRT$^*$                      & 35/50 & 24/50 & 0.663 & 10/50 & 8/50  & 0.180 \\
Informed RRT$^*$             & 33/50 & 24/50 & 0.650 & 13/50 & 10/50 & 0.248 \\
\acs{ESDF} (ground-constr.)  & 38/50 & 26/50 & 0.751 & 12/50 & 8/50  & 0.218 \\
\midrule
Ours (uniform)               & 38/50 & 26/50 & 0.752
                             & 38/50 & 30/50 & 0.694 \\
\quad + refinement           & 40/50 & \secondbest 31/50 & 0.781
                             & \secondbest 39/50 & \secondbest 33/50 & 0.717 \\
Ours (adaptive)              & \secondbest 46/50 & \secondbest 31/50 & \secondbest 0.917
                             & \best 40/50 & 32/50 & \secondbest 0.765 \\
\quad + refinement           & \best 48/50 & \best 34/50 & \best 0.956
                             & \best 40/50 & \best 34/50 & \best 0.782 \\
\bottomrule
\end{tabular*}

\vspace{2pt}
\hfill
{\scriptsize
\setlength{\fboxsep}{1.5pt}
\colorbox{oursgreen!55}{\strut Best}
\hspace{0.6em}
\colorbox{oursgreen!18}{\strut Second best}
}

\end{table}

%% file: fig_quantitative.tex
\begin{figure*}[!t]
	\captionsetup{font=footnotesize}
	\centering
	\includegraphics[width=\textwidth]{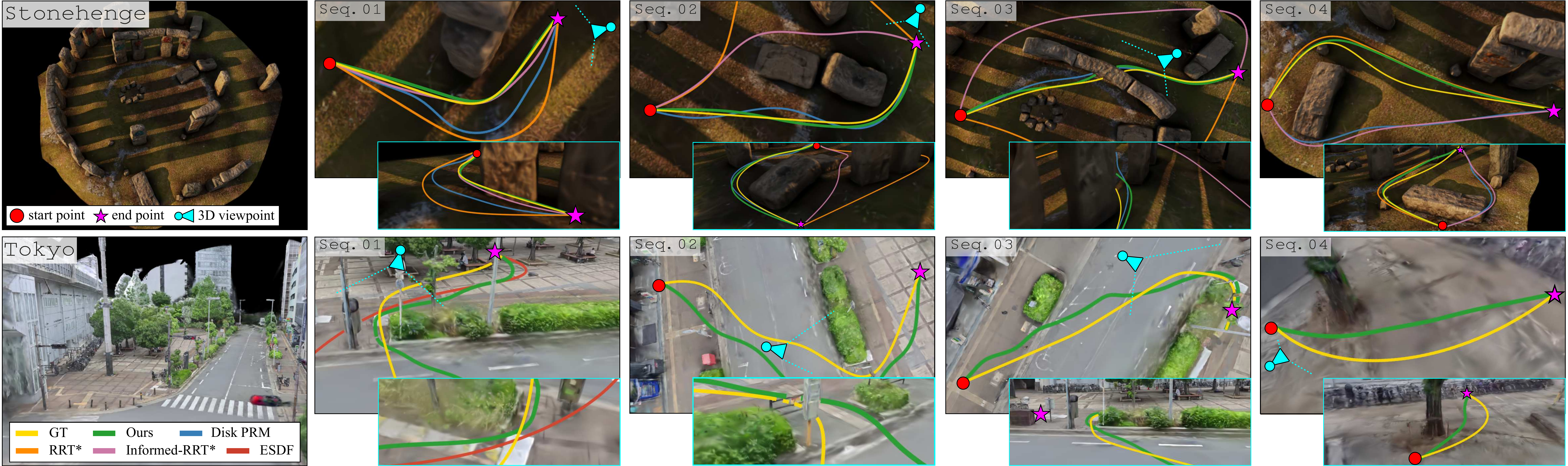}
    \caption{
	Qualitative planning results on \scene{Stonehenge} (top) and \scene{Tokyo} (bottom) for four representative start–goal pairs from Table~\ref{tab:planning}. Each panel overlays the evaluation reference path in yellow with the paths returned by the compared planners; missing trajectories indicate planning failures. Circles and stars denote the start and goal, respectively. For both scenes, the camera symbol indicates the secondary viewpoint shown in the inset. Across these examples, \varTitle{} more consistently returns feasible paths, and its trajectories tend to follow the reference path more closely than those of the compared methods.
	}
	\label{fig:exp_quantitative}
	\vspace{-0.2cm}
\end{figure*}

%% file: d.planning.tex
\subsection{End-to-End Planning}
\label{sec:exp_planning}

We evaluate end-to-end planning on $50$ start--goal pairs per scene in \scene{Stonehenge} and \scene{Tokyo}.
For each start--goal pair, we construct an evaluation-only reference
path using scene-specific reference geometry independently of the
planner.
The reference paths are used only for evaluation.
Table~\ref{tab:planning} reports reference collision-pass counts and \ac{SPL}~\cite{anderson2018evaluation}, while Fig.~\ref{fig:exp_quantitative} shows representative paths.

\paragraph{Planning performance}
\varTitle{} achieves the highest \ac{SPL} on both scenes.
On \scene{Stonehenge}, its collision-pass counts are comparable to those of the largest Disk \ac{PRM} roadmap, while \ac{SPL} is higher.
On \scene{Tokyo}, it improves both collision-pass counts and \ac{SPL} over the primitive-based and ground-constrained \ac{ESDF} baselines.

\paragraph{Node budget and construction cost}
Fig.~\ref{fig:sweep_cost} compares uniform Disk \ac{PRM} across different node budgets.
Larger roadmaps consistently increase construction cost, while the gains in \ac{SPL} diminish on \scene{Stonehenge} and saturate on \scene{Tokyo}.
The strict collision-pass rate likewise does not improve consistently with additional nodes.
The full planner achieves higher \ac{SPL} with lower construction time than the largest Disk \ac{PRM} roadmap on both scenes, showing that simply increasing the uniform sampling budget does not close the performance gap within the tested range.

\paragraph{Effect of \acs{LCC}-based refinement}
\emph{Ours (no refinement)} uses the same roadmap and edge validation as the full method but omits path refinement.
Refinement improves \ac{SPL} on both scenes, while increasing collision-pass counts on \scene{Stonehenge} and maintaining comparable counts on \scene{Tokyo}.
This indicates that the cached \ac{LCC} fields can improve the graph route through refinement without modifying the underlying roadmap.

\paragraph{Online planning latency}
Fig.~\ref{fig:cost} reports the mean per-request latency, including failed queries, separately from offline construction.
Path refinement increases online computation, but the full planner remains faster than RRT$^*$ on both scenes.
On \scene{Tokyo}, its latency is higher than Disk \ac{PRM} and \ac{ESDF}, reflecting the additional computation required for refinement alongside the improved \ac{SPL} and reference collision-pass performance.

%% file: fig_sweep_cost.tex
\begin{figure}[tb]
	\captionsetup{font=footnotesize}
	\centering
	\includegraphics[width=1.0\columnwidth]{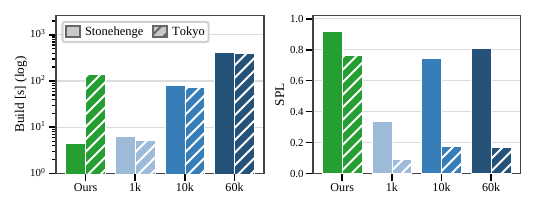}
    \caption{
    Effect of roadmap node budget on construction cost and planning performance. Roadmap build time (left) and SPL (right) are shown for uniform Disk PRM with 1k, 10k, and 60k nodes and for \varTitle{}. \scene{Stonehenge} and \scene{Tokyo} are indicated by solid and hatched bars, respectively. Increasing the uniform sampling budget substantially increases construction time while providing diminishing or saturated SPL gains.
    }
	\label{fig:sweep_cost}
	\vspace{-0.35cm}
\end{figure}

%% file: fig_cost.tex
\begin{figure}[tb]
	\captionsetup{font=footnotesize}
	\centering
	\includegraphics[width=1.0\columnwidth]{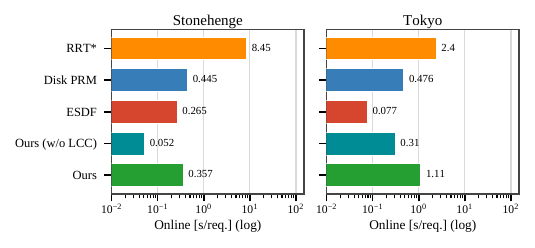}
    \caption{
    Per-request online planning latency for \scene{Stonehenge} and
    \scene{Tokyo}, averaged over $50$ requests including failures and reported
    separately from offline roadmap construction.
    LCC-based path refinement increases online computation, while the full
    \varTitle{} remains faster than RRT$^*$ on both scenes.
    Colors follow the method palette.
    }
	\label{fig:cost}
	\vspace{-0.35cm}
\end{figure}

%% file: fig_robot_size.tex
\begin{figure}[tb]
	\captionsetup{font=footnotesize}
	\centering
	\includegraphics[width=1.0\columnwidth]{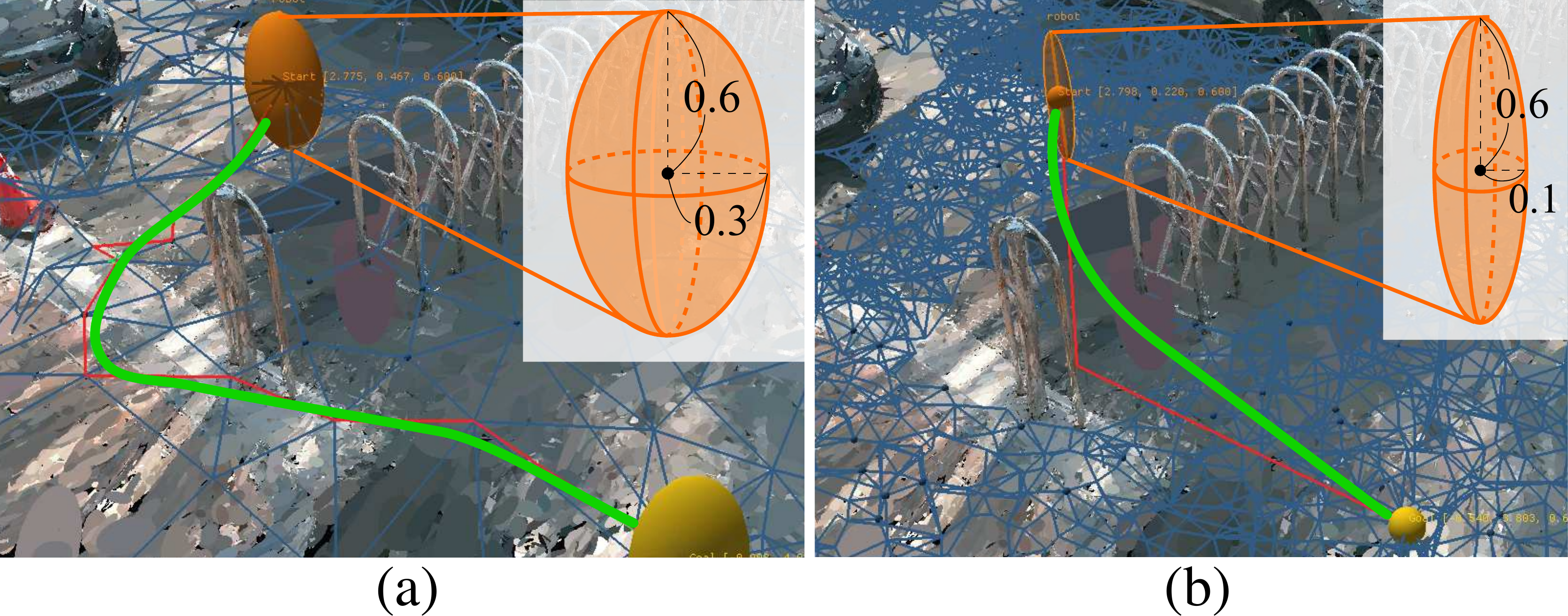}
    \caption{
Robot-size-aware planning through footprint-specific edge validation. (a) Initial graph route (red), refined path (green), and robot body with lateral half-width \(b=0.3\,\mathrm{m}\); the narrower passage is infeasible, leading to the wider route. (b) With \(b=0.1\,\mathrm{m}\), the narrower passage becomes feasible and yields a shorter path. Candidate edges and the refined path are validated using the corresponding orthographic robot footprint.    }
	\label{fig:qual_robot_size}
    \vspace{-0.2cm}
\end{figure}

%% file: f.qualitative.tex
\subsection{Robot-Size-Aware Planning}
\label{sec:exp_robot_size}

Fig.~\ref{fig:qual_robot_size} shows the effect of robot size on planning.
A robot with a lateral half-width of $0.3$ is routed through
the wider passage, while reducing the half-width to $0.1$
allows the planner to use the narrower passage and obtain a shorter path.
In both cases, the refined path is revalidated with the corresponding
robot-specific orthographic footprint query.

%% file: e.real.tex
\subsection{Real-World Humanoid Deployment}
\label{sec:exp_real}

We demonstrate \varTitle{} on a humanoid robot in a real-world environment (Fig.~\ref{fig:real_world}).
The reconstructed \ac{2DGS} map is registered with the LiDAR map from FAST-LIO2~\cite{xu2022fastlio2}, allowing the planned path to be transferred to the robot.
The robot tracks the path using a pure-pursuit controller, demonstrating successful physical execution.

%% file: fig_real.tex
\begin{figure}[t]
	\captionsetup{font=footnotesize}
	\centering
	\includegraphics[width=1.0\columnwidth]{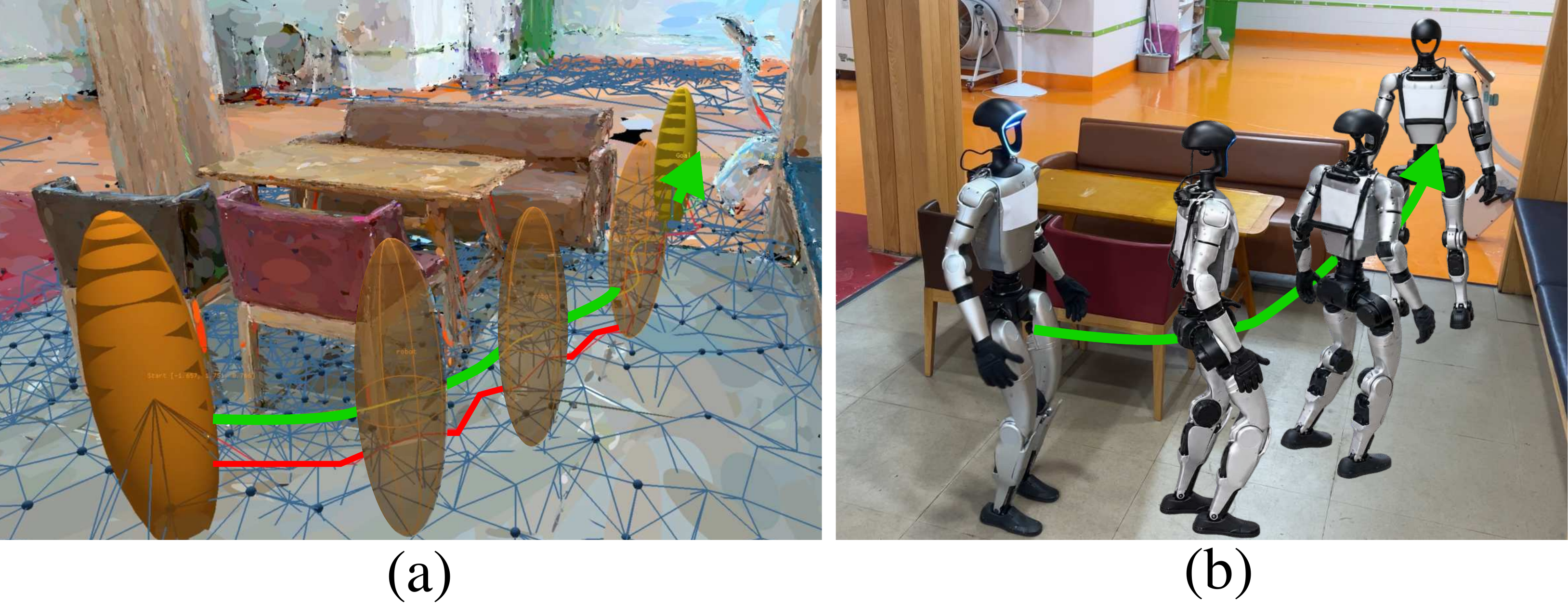}
    \caption{
Real-world deployment of \varTitle{} on a humanoid robot. (a) Initial graph route (red), refined path (green), and corresponding robot body poses in the reconstructed 2DGS map. (b) Autonomous execution of the refined path in the corresponding real-world environment.    }
	\label{fig:real_world}
    \vspace{-0.2cm}
\end{figure}

%% file: conclusion.tex
\section{Conclusion}\label{sec:conclusion}

We proposed \varTitle{}, a path planning framework that directly queries planning-relevant geometry from \ac{2DGS} maps through rasterization.
Experiments on public and self-collected scenes demonstrated the effectiveness of the proposed approach for clearance estimation and end-to-end path planning, with improved performance over the evaluated baselines.
These results suggest that rasterization provides a practical interface for planning directly on Gaussian scene representations.
The current formulation approximates the robot body with an ellipsoid, which limits the representation of detailed and articulated robot geometry, and does not explicitly account for robot dynamics.
Future work will consider more detailed robot-body models and incorporate kinematic and dynamic feasibility into the planning process.